\documentclass{article}
\usepackage[preprint]{spconf}
\usepackage{amsmath,graphicx,amsfonts, amssymb}
\usepackage{arydshln}
\usepackage{xcolor}
\usepackage{lipsum}
\usepackage{microtype}
\usepackage{graphicx}
\usepackage{subfigure}
\usepackage[hidelinks]{hyperref}       
\usepackage{booktabs} 
\usepackage{amsfonts}       
\usepackage{nicefrac}       
\usepackage{arydshln}
\usepackage{algpseudocode}
\usepackage{algorithm}

\usepackage[
backend=biber,
style=ieee,
citestyle=numeric-comp,
maxbibnames=3,
maxcitenames=3,
doi=false,isbn=false,url=false,eprint=false
]{biblatex}
\usepackage{multicol}
\usepackage{multirow}

\defbibheading{bibliography}[\refname]{}
\DeclareSourcemap{
	\maps[datatype=bibtex, overwrite=true]{
		\map{
			\step[fieldsource=booktitle,
			match=\regexp{.*Interspeech.*},
			replace={Proc. Interspeech}]
			\step[fieldsource=journal,
			match=\regexp{.*INTERSPEECH.*},
			replace={Proc. Interspeech}]
			\step[fieldsource=booktitle,
			match=\regexp{.*ICASSP.*},
			replace={Proc. ICASSP}]
			\step[fieldsource=booktitle,
			match=\regexp{.*icassp_inpress.*},
			replace={Proc. ICASSP (in press)}]
			\step[fieldsource=booktitle,
			match=\regexp{.*Acoustics,.*Speech.*and.*Signal.*Processing.*},
			replace={Proc. ICASSP}]
			\step[fieldsource=booktitle,
			match=\regexp{.*International.*Conference.*on.*Learning.*Representations.*},
			replace={Proc. ICLR}]
			\step[fieldsource=booktitle,
			match=\regexp{.*International.*Conference.*on.*Computational.*Linguistics.*},
			replace={Proc. COLING}]
			\step[fieldsource=booktitle,
			match=\regexp{.*SIGdial.*Meeting.*on.*Discourse.*and.*Dialogue.*},
			replace={Proc. SIGDIAL}]
			\step[fieldsource=booktitle,
			match=\regexp{.*International.*Conference.*on.*Machine.*Learning.*},
			replace={Proc. ICML}]
			\step[fieldsource=booktitle,
			match=\regexp{.*North.*American.*Chapter.*of.*the.*Association.*for.*Computational.*Linguistics:.*Human.*Language.*Technologies.*},
			replace={Proc. NAACL}]
			\step[fieldsource=booktitle,
			match=\regexp{.*Empirical.*Methods.*in.*Natural.*Language.*Processing.*},
			replace={Proc. EMNLP}]
			\step[fieldsource=booktitle,
			match=\regexp{.*Association.*for.*Computational.*Linguistics.*},
			replace={Proc. ACL}]
			\step[fieldsource=booktitle,
			match=\regexp{.*Automatic.*Speech.*Recognition.*and.*Understanding.*},
			replace={Proc. ASRU}]
			\step[fieldsource=booktitle,
			match=\regexp{.*Spoken.*Language.*Technology.*},
			replace={Proc. SLT}]
			\step[fieldsource=booktitle,
			match=\regexp{.*Speech.*Synthesis.*Workshop.*},
			replace={Proc. SSW}]
			\step[fieldsource=booktitle,
			match=\regexp{.*workshop.*on.*speech.*synthesis.*},
			replace={Proc. SSW}]
			\step[fieldsource=booktitle,
			match=\regexp{.*Advances.*in.*neural.*information.*processing.*},
			replace={Proc. NeurIPS}]
			\step[fieldsource=booktitle,
			match=\regexp{.*Advances.*in.*Neural.*Information.*Processing.*},
			replace={Proc. NeurIPS}]
			\step[fieldsource=booktitle,
			match=\regexp{.*Workshop.*on.* Applications.* of.* Signal.*Processing.*to.*Audio.*and.*Acoustics.*},
			replace={Proc. WASPAA}]
			\step[fieldsource=publisher,
			match=\regexp{.+},
			replace={{}}]
			\step[fieldsource=month,
			match=\regexp{.+},
			replace={{}}]
			\step[fieldsource=location,
			match=\regexp{.+},
			replace={{}}]
			\step[fieldsource=address,
			match=\regexp{.+},
			replace={{}}]
			\step[fieldsource=organization,
			match=\regexp{.+},
			replace={{}}]
		}
	}
}
\usepackage{cancel}

\makeatletter
\def\blfootnote{\xdef\@thefnmark{}\@footnotetext}
\makeatother
\usepackage{multicol}
\usepackage{multirow}

\title{Joint Prediction and Denoising for Large-Scale Multilingual Self-Supervised Learning}
\name{
    \begin{tabular}[c]{@{}c@{}c@{}c@{}c@{}c@{}}
        William Chen$^1$, Jiatong Shi$^1$, Brian Yan$^1$, Dan Berrebbi$^1$, Wangyou Zhang$^{1,2}$, Yifan Peng$^1$, \\ Xuankai Chang$^1$, Soumi Maiti$^1$, Shinji Watanabe$^1$
    \end{tabular}
}

\address{$^1$Carnegie Mellon University, USA\quad
$^2$Shanghai Jiao Tong University, China
}
\copyrightnotice{\copyright\ IEEE 2023}
\toappear{To appear in {\it Proc.\ ASRU2023, December 16-20, 2023, Beitou, Taipei}}
\begin{document}
%
\maketitle

\begin{abstract}
Multilingual self-supervised learning (SSL) has often lagged behind state-of-the-art (SOTA) methods due to the expenses and complexity required to handle many languages. This further harms the reproducibility of SSL, which is already limited to few research groups due to its resource usage. We show that more powerful techniques can actually lead to more efficient pre-training, opening SSL to more research groups. We propose WavLabLM, which extends WavLM's joint prediction and denoising to 40k hours of data across 136 languages. To build WavLabLM, we devise a novel multi-stage pre-training method, designed to address the language imbalance of multilingual data. WavLabLM achieves comparable performance to XLS-R on ML-SUPERB with less than 10\% of the training data, making SSL realizable with academic compute. We show that further efficiency can be achieved with a vanilla HuBERT Base model, which can maintain 94\% of XLS-R’s performance with only 3\% of the data, 4 GPUs, and limited trials. We open-source all code and models in ESPnet.

\end{abstract}
\vspace{-0.3cm}
\section{Introduction}
\vspace{-0.3cm}
\label{sec:intro}

Recent advances in NLP and speech processing have been driven by large scale self-supervised learning (SSL) models trained on gigantic amounts of unlabeled or weakly labeled data \cite{radford2022robust, brownGpt, ChenWavLm, sslReview, sslASRReview, brownGpt, radford2022robust, chowdhery2022palm, scao2022bloom, black2022gptneoxb, srivastava2022beyond}. In speech processing specifically, SSL research has largely centered around speech representation learning, accomplished by training large speech encoders \cite{schneider19_wav2vec,baevskiw2v, hsuHubert, ChenWavLm, wangilsssl} on tens of thousands of hours of unlabeled audio. 

By leveraging the abundant amounts of unlabeled data, SSL presents a scalable solution for extending the reach of speech technologies to new audiences. This can be most directly accomplished via cross-lingual transfer learning, which aims to leverage information from other languages within a single model. However, due to the increased complexity and computational expenses of training with multiple languages, the methodologies used in multilingual SSL have often lagged behind the state-of-the-art, which remain centered around high-resource languages such as English \cite{ChenWavLm, chungW2vBert, radford2022robust, sslReview, pmlr-v162-baevski22a}. 

This can be evidenced via the SUPERB Benchmark \cite{yang21c_interspeech}, which aims to evaluate the effectiveness of different SSL models across an array of speech processing tasks for English. While masked-prediction methods such as HuBERT \cite{hsuHubert} and WavLM \cite{ChenWavLm} reign supreme on SUPERB, they are notably absent in multilingual studies \cite{chenFLEURS, rouditchenko2023comparison, khurana2022samu, cahyawijaya2023cross, ashihara2023speechglue, sikasote2023zambezi}. Instead, popular multilingual SSL models such as XLSR 53 \cite{conneau2020unsupervised}, XLS-R 128 \cite{babu2021xls}, and MMS \cite{pratap2023scaling} are built upon the wav2vec framework \cite{schneider19_wav2vec, baevskiw2v}\footnote{A mHuBERT \cite{lee-etal-2022-textless} model was publicly released, but was only pre-trained on 3 languages: English, Spanish, and French, and on a much smaller scale.}, which is outperformed by the mentioned masked-prediction models above \cite{pmlr-v162-chiu22a, pmlr-v162-baevski22a, hsuHubert,ChenWavLm,zaiem2023speech}.

Our goal is to build a large-scale multilingual SSL model that uses state-of-the-art (SOTA) SSL methods while maintaining a reasonable computational budget. We believe that using \textit{more powerful techniques} can actually lead to\textit{ more efficient pre-training}, as less data and compute is needed to reach similar levels of performance. By showing the feasibility of large-scale multilingual SSL without industrial compute, we hope to encourage more community-driven efforts in 
building state-of-the-art multilingual models and help pave the way for extending speech processing to new languages \cite{shi2023ml, scao2022bloom}.

As such, we propose a new massively multilingual SSL model: WavLabLM. WavLabLM extends the WavLM framework of jointly learning masked speech prediction and denoising to nearly 40k hours of speech across 136 languages. Despite its relatively small dataset size \cite{conneau2020unsupervised, zhang2023google, babu2021xls}, WavLabLM's pre-training data is curated for wide geographical coverage, the 15 languages with the most data span 7 different regions across the globe. For contrast, the entire top 15 languages used for XLS-R 128 \cite{babu2021xls} are European.

To pre-train WavLabLM within academic constraints, we explore various techniques to make SSL more efficient. We first experimented with continual learning from English-based SSL models, halving the resources required for pre-training at the cost of performance in certain tasks. For multilingual SSL models trained from scratch, we propose a novel multi-stage pre-training approach to counteract imbalances in the language distribution of the data. This can be done by simply training on a large unbalanced dataset, before continual learning on a smaller dataset with a balanced language distribution. As a result, WavLabLM achieves comparable performance to XLS-R 128 on the 10-minute track of ML-SUPERB, despite only using 10\% of the pre-training data, and even outperforms it in several tasks. 

To better fit SSL pre-training into academic constraints, we further slim down our approach.
Using only 4 GPUs, we find that a simple HuBERT Base model is surprisingly efficient at multilingual SSL. Our CV-HuBERT, trained only on 92 languages from Common Voice \cite{ardila-etal-2020-common}, maintains up to 94\% of the performance of XLS-R with only 3\% of the training data and a third of the model parameters. In summary, our methods and results can be outlined as follows:
\begin{itemize}
    \item We propose WavLabLM, an SSL model for 136 languages and the first extension of joint denoising and prediction to multilingual SSL.
    \item We propose novel SSL techniques to make pre-training more efficient: continual pre-training from English models and multi-stage pre-training for language balancing.
    \item We explore multi-resolution inputs in the multilingual setting, and show significant efficiency gains can be achieved with only a vanilla HuBERT Base model. 
    \item Overall, WavLabLM achieves comparable performance to XLS-R 128 with only 10\% of the pre-training data. Our CV-HuBERT maintains up to 94\% of the performance with only 3\% of the data.
\end{itemize}
Finally, we include detailed descriptions of our engineering work that was required for scaling SSL, as a community contribution. We open-source all code and trained models in the ESPNet toolkit \cite{watanabe2018espnet}.

\vspace{-0.3cm}
\section{Background}
\vspace{-0.3cm}
\subsection{HuBERT} \label{sec:hubert}
\vspace{-0.2cm}
WavLabLM uses masked-prediction based pre-training (MPPT), which has become a popular pre-training objective among SOTA SSL models \cite{hsuHubert, chungW2vBert, pmlr-v162-chiu22a, ChenWavLm, baevskiw2v}. MPPT models are trained by predicting discretized representations of masked regions of the input speech. In doing so, the model is forced to leverage the unmasked audio as context for its predictions.

MPTT models largely differ in how the discrete representations are obtained. Wav2vec 2.0 \cite{baevskiw2v} and w2v-BERT \cite{chungW2vBert} learn the discrete units as part of the SSL training, allowing the representations to also be trained end-to-end. While powerful, these models are prone to codebook collapse during training \cite{zhang2023google}, and require an additional loss to keep the discrete units diverse. We instead adopt the offline clustering approach of HuBERT \cite{hsuHubert}, which has been successfully been adapted to other SSL models such as WavLM \cite{ChenWavLm}. HuBERT takes an iterative approach to SSL through an offline clustering step: prior to pre-training, the input data is clustered using k-means. HuBERT is thus trained to predict the frame-level cluster assignments of masked regions of the input speech. 

\vspace{-0.3cm}
\subsection{WavLM} \label{sec:wavlm}
\vspace{-0.2cm}
While many SSL models are trained on clean audiobook read-speech \cite{baevskiw2v, hsuHubert, schneider19_wav2vec}, real world scenarios often feature noisy and multi-speaker settings. To address this, WavLM \cite{ChenWavLm} jointly trains the model to perform both masked speech prediction and denoising. This is done by augmenting the input speech via dynamic mixing with either noise or overlapping speech. 

For every utterance $u$ in an input batch $B$, a random value $v$ is sampled from the continuous distribution  $\mathcal{U}(0,1)$. If $v$ is less than the augmentation probability $p_n$, another random value $w$ is sampled from $\mathcal{U}(0,1)$. If $w$ is less than utterance mixing probability $p_u$, another utterance $u_n$ in the batch $B$ is sampled for mixing. Otherwise, a random Deep Noise Suppression (DNS) noise $u_n$ is used \cite{reddy2020interspeech}.

A mixing energy ratio $e$ is then sampled for the mixed audio. If another utterance is used, $e$ is sampled from $\mathcal{U}(-5,20)$. Otherwise, $e$ is sampled from $\mathcal{U}(-5,5)$. The starting timestamp of $u$ for the mixing $t$ is sampled in two stages. First, we sample a minimum length $l_\text{min}$ from the discrete uniform distribution $\mathcal{U}(0,\frac{\text{len}(u)}{2})$. This way, $t$ can be sampled from $\mathcal{U}(l_\text{min},\text{len}(u_n))$. The same process is applied again to obtain $t_n$, the starting timestamp for mixing in $u_n$. Next, the energy of the utterances is calculated as $E_u = \frac{\sum u * u}{\text{len(u)}}$ and $E_n  = \frac{\sum u_n * u_n}{\text{len(u$_n$)}}$ for $u$ and $u_n$ respectively. Then, the mixing scale $s$ is calculated as $s = \sqrt{\frac{E_u}{e * E_n}}$. Finally, the portion where the augmentation is applied $u[t : t + l]$ is summed with the scaled noise portion $s * u_n[t_n : t_n + l]$.

By applying this augmentation during training, WavLM is made more robust to noisy and multi-speaker settings, and achieves SOTA results on the SUPERB Benchmark \cite{yang21c_interspeech}.

\vspace{-0.3cm}
\section{Pre-training WavLabLM} \label{sec:WavLabLM}
\vspace{-0.3cm}

\subsection{Pre-training Data} \label{sec:data}
\vspace{-0.3cm}
\begin{table}[]
    \centering
        \caption{List of corpora in our pre-training data. For VoxPopuli, we only use recordings from the annotated subset (but discard the transcriptions for unsupervised SSL).}
    \label{tab:data}

    \begin{tabular}{l|c|c|c}
        \toprule
        Dataset & Languages & Hours & Speech Type\\
\midrule
 MLS & 8 & 22,185 & Read\\ 
 Common Voice & 92 & 13,600 & Read\\ 
 Googlei18n & 34 & 1,328 & Mixed\\ 
 VoxPopuli & 16 & 1,024 & Spontaneous\\ 
 BABEL & 17 & 1000 & Spontaneous\\ 
 FLEURS & 102 & 950 & Read\\
 \midrule
 Total & 136 & 39,087\\
\bottomrule
    \end{tabular}

    \vspace{-0.6cm}
\end{table}
To conduct pre-training, we build Openli110, a large-scale multilingual training corpus that covers 109 languages. It consists of  Common Voice\cite{ardila-etal-2020-common}, VoxPopuli \cite{wang-etal-2021-voxpopuli}, MLS \cite{pratap20_interspeech_mls}, and Googlei18n \footnote{ Resources 32, 35, 36, 37, 41, 42, 43, 44, 52, 53, 54, 61, 63, 64, 65, 66, 69, 70, 71, 72, 73, 74, 75, 76, 77, 78, 79, and 86 from \url{openslr.org} }. To extend coverage to 136 languages, we also include BABEL \cite{gales2014speech} and FLEURS \cite{conneau2022fleurs}. A full break down  of our pre-training data by corpus is available in Table \ref{tab:data}. We filter out audio clips longer than 15 seconds due to GPU memory constraints, leaving us with 40k hours of pre-training data. We want to emphasize the geographical diversity of this data: the top 15 languages are spread across 7 different regions across the globe (Table \ref{tab:langs}). In contrast, the entire top 15 languages used for the SOTA XLS-R \cite{babu2021xls} are West/East European. We use the FLEURS development split for a balanced validation, which has similar amounts of data per language.

\begin{table}[tb]
    \centering
        \caption{The top 15 languages in our SSL pre-training set by geographic region and hours. Together, they account for 92.5\% of the data. }
    \label{tab:langs}

    \begin{tabular}{l|c|c}
        \toprule
        Language & Region & Hours\\
\midrule
 English & West Europe & 23663 \\
 Kinyarwanda & Sub-Saharan Africa & 1984\\
 Esperanto & Constructed & 1360\\
 German & West Europe & 1256\\
 Catalan & West Europe & 1184\\
 Belarussian & East Europe & 965\\
 French & West Europe & 934\\
 Spanish & West Europe &523\\
 Kabyle & North Africa & 518\\
 Ganda & Sub-Saharan Africa & 362\\
 Sundanese & Southeast Asia &  330\\
 Italian & West Europe & 314\\
 Javanese & Southeast Asia & 294\\
 Bengali & South Asia & 239\\
 Sinhala & South Asia & 218\\
 Bashkir & Central Asia & 216\\
 \midrule
 Total & & 34362 \\
\bottomrule
    \end{tabular}

    \vspace{-0.5cm}
\end{table}

  \vspace{-0.3cm}
\subsection{Pre-Training Architecture} \label{sec:pretraining-target}
\vspace{-0.2cm}
\begin{figure*}[]
    \centering
    \includegraphics[height=5cm]{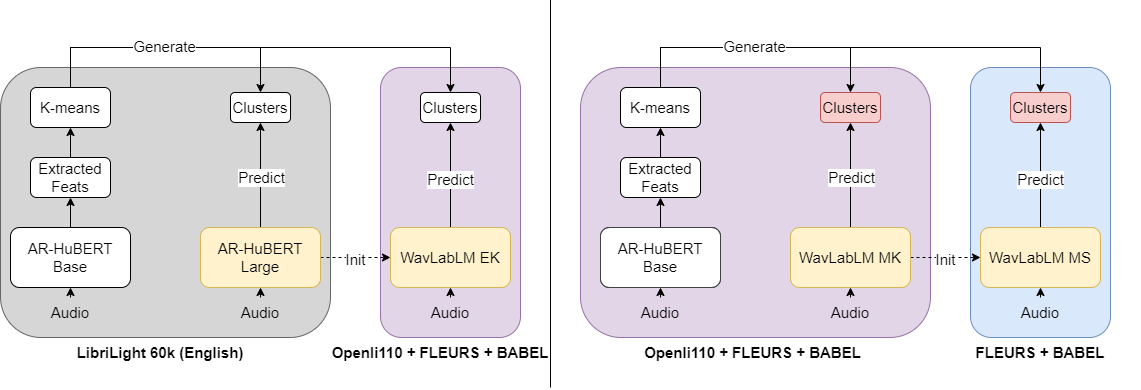}
    \caption{Diagram of our proposed approaches. WavLabLM EK is initialized with HuBERT Large and uses English-based k-means (left). WavLabLM MK is trained from scratch with multilingual k-means, and used to initialize WavLabLM MS (right).}
    \label{fig:arch-fig}
    \vspace{-0.3cm}
\end{figure*}

WavLabLM uses the HuBERT \cite{hsuHubert} architecture, which modified wav2vec 2.0 \cite{baevskiw2v} for codebook prediction. It consists of a CNN feature extractor, a Transformer \cite{vaswani2017attention} encoder, and a codebook output layer. WavLabLM also uses the HuBERT pre-training objective detailed in Section \ref{sec:hubert}; it predicts the cluster assignments of each masked audio frame. One major novelty of our work is to enhance model robustness to noise in multilingual settings by applying the WavLM joint denoising task discussed in Section \ref{sec:wavlm}.  We do not adopt the architectural changes introduced by WavLM, such as gated relative positional bias and scaled softmax. This is because it prevents the application of the second novelty of our method, continual learning from existing HuBERT models, as described below.

The simplicity of the HuBERT approach to SSL allows it to be both scalable and versatile. Since the clusters are obtained offline, we can reduce resource consumption by using existing SSL models for feature extraction. To show that powerful multilingual SSL is feasible without large-scale compute, we adopt AR-HuBERT Base, an English HuBERT Base model replicated through academically-resourced SSL \cite{chen2023reducing} \footnote{Improved performance can be likely achieved by using existing multilingual SSL models, such as XLS-R \cite{babu2021xls}, for feature extraction. For HuBERT SSL, feature quality is a key indicator of downstream performance \cite{chen2023reducing, hsuHubert}}. We can then use AR-HuBERT Base to extract features from our multilingual pre-training dataset and train a multilingual k-means. Since we are using AR-HuBERT Base, we can also keep following the HuBERT pipeline and recycle the k-means model used for AR-HuBERT Large, another academic reproduction \cite{chen2023reducing}. As such, we can perform continual learning on AR-HuBERT Large to adapt it to a multilingual setting (Figure \ref{fig:arch-fig}, left), which we denote as \textit{WavLabLM EK}. While the multilingual k-means is likely more optimal since the discrete representations will be more expressive of multilingual speech, continual learning can significantly reduce compute costs while leveraging knowledge from English pre-training.

\vspace{-0.3cm}
\subsection{Multi-stage Pre-training} \label{sec:multi}
\vspace{-0.2cm}
A persistent challenge in training multilingual models is the data imbalance between languages. For example, over half of our the data consists of English speech (Table \ref{tab:langs}), which is not uncommon in these large-scale settings. A common way to address this issue is through temperature-based upsampling, where speech from lower-resource languages has an increased chance of being sampled into a batch \cite{babu2021xls, pratap2023scaling}. While effective, this is difficult to scale to large collections of multiple corpora: different datasets use different language labelling standards, if labels are provided at all. Furthermore, it also suffers from increased computational overhead and relies on additional tuning of the temperature parameter.

To reduce the experimental mass while obtaining balanced performance by languages, we instead adopt a simple two-stage training approach (Figure \ref{fig:arch-fig}, right). We first pre-train on a large unbalanced portion of the data, before performing continual learning on a smaller balanced subset. We first pre-train only on the full 40k hours described in Section \ref{sec:data}, which is split unevenly across 136 languages (Table \ref{tab:langs}), to obtain \textit{WavLabLM MK}. We then continue pre-training on a subset consisting of FLEURS and BABEL (total 2000 hours), yielding \textit{WavLabLM MS}. We chose FLEURS due to its equal language spread across 102 languages, and BABEL to retain a diverse selection of conversational speech.

\vspace{-0.3cm}
\subsection{Pre-training Settings}
\vspace{-0.2cm}
WavLabLM uses the HuBERT Large architecture \cite{hsuHubert}. We follow the same hyperparameters as Chen. et al. \cite{chen2023reducing} in their reproduction. The model consists of 24 Transformer \cite{vaswani2017attention} encoder layers, each with a hidden size of 1024, a feed-forward dimension of 4096, and an output feature size of 768. The model uses the Adam optimizer \cite{kingma2014adam} with 32,000 warmup steps and a peak learning rate of 0.0005. Models were pre-trained on 32 40GB Nvidia A100 GPUs \footnote{Although we obtained such large GPU resources from public computing centers, our occupation of these GPU resources was very limited. We could only perform a few experimental trials, and thus our study still remains within an academic computing scale. In comparison, XLS-R 128 used 128 GPUs, a much larger computing scale.}, and each GPU handled at most 30 seconds of audio per batch. As discussed in Section \ref{sec:wavlm}, we use an augmentation probability $p_n$of 0.2 and a utterance mixing probability $p_u$ of 0.1. The energy ratio of a DNS noise is sampled from a continuous uniform distribution of -5 to 5, while the energy ratio of a mixed utterance is sampled from -5 to 20. For the multi-stage model described in Section \ref{sec:multi}, we perform pre-training on the FLEURS+BABEL subset for an additional 10k steps.

\vspace{-0.3cm}
\section{Multi-resolution Multilingual SSL} \label{sec:multires}
\vspace{-0.2cm}
While we obtained promising results with our WavLabLM models, the computational resources required to pre-train them were still beyond the typical amounts available to academic groups. As such, we sought to further slim down our approach. We first propose using a smaller model, the 95M parameter HuBERT Base \cite{hsuHubert} architecture, which has been found to occasionally outperform larger variants in monolingual SSL \cite{zaiem2023speech, shi2023ml}. For pre-training data, we use the Common Voice \cite{ardila-etal-2020-common} subset of the Openli110 dataset discussed in Section \ref{sec:data}, which contains 13k hours of data split across 92 languages. We choose Common Voice due to its diverse selection of languages and relatively large size, while still being manageable with fewer resources. We designate this approach as \textit{CV-HuBERT}. To our knowledge, CV-HuBERT is the first attempt in multilingual SSL to match the performance of large-scale multilingual SSL models with smaller ones that use less data.

For further potential efficiency gains, we investigate the effectiveness of SSL at different resolutions, which has shown to be beneficial in monolingual SSL \cite{shi2023exploration} and other speech processing tasks \cite{kimSqueeze, burchiEfficient, zhaoUnet}. As such, we extend the multi-resolution (MR) HuBERT to the multilingual setting. This is first accomplished by training HuBERT models at lower resolutions (40ms and 80ms) than the original architecture (20ms). After pre-training, the models are then fused by stacking the representations from the various resolutions. As such, we pre-train four variations of CV-HuBERT at 20ms, 40ms, and 80ms resolutions. All the models are based on HuBERT-base configuration, except for the convolution feature extractor used for modulating the modeling resolution. For the multi-resolution implementation, we utilize the same parallel architecture discussed in \cite{shi2023exploration}, which repeats the features of low resolution to match the temporal dimension of high-resolution features. All models are trained from scratch, and use the English k-means from AR-HuBERT Base discussed in Section \ref{sec:pretraining-target}. All models were trained with only 4 A100 GPUs. Detailed configurations for each model are shown in Table~\ref{tab:models}.

\vspace{-0.3cm}
\section{Experimental Setup}
\vspace{-0.3cm}
\begin{table*}[htb]
    \centering
    \caption{Results on the ML-SUPERB \texttt{\{10min/1h\}} settings, presented in accuracy (ACC $\uparrow$), character error rate (CER $\downarrow$), and SUPERB score (SUPERB$_\textit{s}$ $\uparrow$). Results of prior SSL models on each task are obtained from the ML-SUPERB Benchmark \cite{shi2023ml}, while SUPERB scores are re-calculated by considering our models.}
    \resizebox {\linewidth} {!} {
\begin{tabular}{lc|c|c|cc|c|ccc}
\toprule
\multirow{3}{*}{} & \multirow{3}{*}{} & \multirow{3}{*}{} & \textsc{Mono. ASR} & \multicolumn{2}{c|}{\textsc{Multi. ASR}} & \multicolumn{1}{c|}{\textsc{LID}} & \multicolumn{3}{c}{\textsc{Multi. ASR + LID}} \\
\cmidrule(lr){4-4} \cmidrule(lr){5-6} \cmidrule(lr){7-7} \cmidrule(lr){8-10}
&    &      &     &        Normal & Few-shot & Normal & \multicolumn{2}{c}{Normal} & \multicolumn{1}{c}{Few-shot} \\
Model & Hours & SUPERB$_{s}$ & CER & CER & CER & ACC & ACC & CER & \multicolumn{1}{c}{CER}  \\
\midrule
XLSR 53 \cite{conneau2020unsupervised} & 56k & 530 / 889 & 49.5 / 34.9 & 33.9 / 26.9 & 43.6 / 40.6 & 6.6 / 87.1 & 45.6 / 76.9 & 33.4 / 28.6 & 43.2 / 44.6  \\
XLS-R 128 \cite{babu2021xls} & 436k & \textbf{917} / \textbf{986} & 39.7 / \textbf{30.6} & \textbf{29.3} / \textbf{22.0} & \textbf{40.9} / \textbf{39.3} & \textbf{66.9} / \textbf{87.9} & 55.6 / 85.6 & \textbf{28.4} / \textbf{22.9} & \textbf{42.1} / 42.4  \\
HuBERT Base \cite{hsuHubert} & 1k  & 806 / 879 & 42.8 / 35.3 & 39.8 / 31.4 & 44.5 / 42.7 & 61.2 / 86.1 & \textbf{71.5} / \textbf{86.0} &  39.2 / 30.9& 43.8 / \textbf{41.8}\\
HuBERT Large \cite{hsuHubert} & 60k & 661 / 778 & \textbf{38.2} / 32.2 & 44.4 / 37.7 & 48.2 / 43.5 & 46.5 / 64.1 & 55.4 / 77.7 & 45.6 / 35.1 & 49.3 / 42.2  \\
\midrule
WavLabLM EK & 40k & 797 / 862 & 40.7 / 33.7 & 41.0 / 33.5 &  44.1 / 41.9 & 61.2 / 83.4 & 60.0 / 79.9 & 40.0 / 33.1 & 42.6 / 41.3 \\
WavLabLM MK & 40k & 845 / 848 & 40.5 / 32.3 & 38.8 / 32.8 & 44.4 / 42.8 & 67.6 / 79.0 & 69.0 / 79.6 & 38.6 / 32.8 & 44.2 / 42.4  \\
WavLabLM MS & 40k & \textbf{908} / \textbf{897} & \textbf{39.9} / \textbf{32.1} & \textbf{37.8} / \textbf{31.3} & \textbf{43.8} / \textbf{40.9} & \textbf{71.7} / 81.1 & \textbf{70.8} / \textbf{82.2} & \textbf{37.0} / \textbf{30.6} & \textbf{43.4} / \textbf{40.2}  \\
\midrule
CV-HuBERT 20ms & 13k & \textbf{892} / \textbf{924} &  \textbf{41.9} / \textbf{32.9} & \textbf{35.4} / \textbf{27.5} & 44.0 / 40.8 & \textbf{71.2} / 84.0 & \textbf{76.6} / \textbf{87.3} & \textbf{35.1} / \textbf{28.2} & 43.6 / 41.1  \\
CV-HuBERT 40ms & 13k & 309 / 409 & 71.6 / 62.6 & 60.5 / 52.0 & 57.5 / 53.0 & 65.6 / 83.0 & 65.7 / 83.3 & 59.6 / 52.3 & 57.7 / 53.4  \\
CV-HuBERT 80ms  & 13k & -127 / -26 & 76.4 / 67.6 & 72.7 / 70.7 & 66.1 / 64.1 & 33.2 / 57.2 & 17.2 / 39.4 & 72.3 / 70.4 & 64.2 / 64.4 \\ 
CV-HuBERT MR  & 13k & 819 / 877 & 47.8 / 38.3 & 37.0 / 28.3 & \textbf{43.2} / 40.8 & 64.1 / \textbf{86.0} & 74.8 / 84.5 & 36.2 / 30.6 & \textbf{42.5} / \textbf{41.0} \\
\bottomrule
\end{tabular}
}
    \label{tab:results_10m}
    \vspace{-0.5cm}
\end{table*}

\begin{table}[t]
    \centering
    \caption{Proposed models evaluated on ML-SUPERB}
    \resizebox {\linewidth} {!} {

    \begin{tabular}{lccccc}
        \toprule
        \multirow{2}{*}{Model} & \multirow{2}{*}{Params}  & \multicolumn{2}{c}{Compute} &  \multicolumn{2}{c}{Data}   \\
         &  & \# GPUs & \# Hours &  \# Langs &  \# Hours  \\
\midrule
WavLabLM EK & 317M & 32 & 3.3k &  136 & 40k \\
WavLabLM MK & 317M & 32 & 6.4k  &  136 & 40k \\
WavLabLM MS & 317M & 32 & 6.5k &  136 & 40k \\
 \midrule
 CV-HuBERT 20ms & 95M & 4 & 1.1k & 92 & 13k  \\
 CV-HuBERT 40ms  & 96M & 4 & 0.9k & 92 & 13k \\
 CV-HuBERT 80ms  & 96M & 4 & 0.5k & 92 & 13k \\
 CV-HuBERT MR & 287M & 4 & 2.5k & 92 & 13k \\
        \bottomrule
    \end{tabular}

}
    \label{tab:models}
    \vspace{-0.6cm}
\end{table}
We conducted all experiments using the ESPnet toolkit \cite{watanabe2018espnet}. We evaluate three variants of WavLabLM, which was proposed in Section \ref{sec:WavLabLM}. We also conduct separate ablations on HuBERT SSL at four different resolutions, as noted in Section \ref{sec:multires}. Table \ref{tab:models} presents details for the data and parameters of each model.

The three variants of WavLabLM are all pre-trained on the 40k hours of multilingual data described in Section \ref{sec:data}. \textit{WavLabLM EK} is initialized from a reproduced HuBERT Large model \cite{chen2023reducing}, and uses an English-based k-means for clustering. \textit{WavLabLM MK} is trained from scratch and uses a multilingual k-means derived from the training data. Finally, \textit{WavLabLM MS} uses the multi-stage training approach described in Section \ref{sec:multi} to balance the language performance of WavLabLM MK. For the resolution experiments, we trial HuBERT Base SSL on Common Voice at 20ms, 40ms, and 80ms. These models are denoted as \textit{CV-HuBERT} \textit{20ms} / \textit{40ms} / \textit{80ms},respectively. As discussed in Sec.~\ref{sec:multires}, we also stack the above HuBERT with different resolutions to construct HuBERT with multiple resolutions, namely \textit{CV-HuBERT MR}. 

Due to the significant computational resources required for each trial (Table \ref{tab:models}), we do not conduct extensive ablations for all possible pre-training cluster / model initialization configurations. We instead focus on the most practical use cases for each technique. For example, the most significant benefit of using English k-means is the initialization of all weights from a pre-trained English model. Without initialization, the multilingual k-means will generally be the better choice, as it better captures the acoustic diversity of the data. 
\vspace{-0.3cm}
\subsection{Evaluation Data and Metrics} \label{sec:eval}
\vspace{-0.2cm}
We conduct extensive evaluations of our models on the Multilingual SUPERB (ML-SUPERB) Benchmark \cite{shi2023ml}. ML-SUPERB consists of several tracks across a 10 minute and 1 hour training data setting:

\begin{itemize}
    \item 14 monolingual ASR tasks split across 9 languages.
    \item A 143-language multilingual ASR task. 20 of the 143 languages have only 5 utterances available for training.
    \item A language identification (LID) task for 143 languages.
    \item A 143-language joint LID+ASR task. 20 of the 143 languages have only 5 utterances available for training.
\end{itemize}
\vspace{-0.1cm}
Models are evaluated using accuracy (ACC) and Character Error Rate (CER). An overall score (SUPERB$_\textit{s}$) is calculated using the same methodology as the SUPERB Benchmark \cite{yang21c_interspeech, shi2023ml} - the score of each model on each task is normalized by the score of the best model, before being averaged across tasks and multiplied by 1000.

\vspace{-0.3cm}
\subsection{Fine-tuning Settings}
\vspace{-0.2cm}
Models are fine-tuned on each track of ML-SUPERB for evaluation. ML-SUPERB follows same style as the SUPERB \cite{yang21c_interspeech} Benchmark: the weights of the SSL model are frozen during training. A learned weighted sum of the layer outputs is fed to a Transformer \cite{vaswani2017attention} encoder trained using CTC loss \cite{graves2006connectionist}. We use the default ML-SUPERB settings for all models, which has 2 encoder layers, each with a hidden dimension of 256, 8 attention heads, and a 1024 feed-forward size. The model is trained with the Adam optimizer \cite{kingma2014adam} with a task-specific constant learning rate. During training, the SSL features are augmented with SpecAug \cite{park2019specaugment}.

\vspace{-0.3cm}
\section{Results}
\vspace{-0.3cm}


Our main results on ML-SUPERB are presented in Table \ref{tab:results_10m}. In each track, we compare against XLSR 53 \cite{conneau2020unsupervised}, XLS-R 128 \cite{babu2021xls}, HuBERT Base \cite{hsuHubert}, and HuBERT Large \cite{hsuHubert}, which are the four of the best performing models on ML-SUPERB \cite{shi2023ml}. As discussed in Section \ref{sec:eval}, models are evaluated using accuracy, CER, and SUPERB Score (SUPERB$_\textit{s}$).

\begin{table}[htb]
    \centering
    \caption{Results by language and pre-training hours on the ML-SUPERB monolingual track, reported in CER for both the \texttt{\{10min/1h\}} settings.}
    \resizebox {\linewidth} {!} {

    \begin{tabular}{lccc}
        \toprule
        Lang. & Hrs & WavLabLM MK & WavLabLM MS\\
\midrule
 English & 24k & \textbf{38.5} / \textbf{31.9 }  & 39.7 / 32.6\\
 French & 940 & \textbf{53.0} / \textbf{39.0} &  53.1 / 39.3\\
 German & 1.3k & \textbf{36.5} / \textbf{28.0} & 38.5 / 28.7 \\
 Russian & 155 & 34.3 / 28.9  & \textbf{34.0} / \textbf{28.7}\\
 Swahili & 160 & 36.1 / 29.2  & \textbf{34.9} / \textbf{27.2}\\
 Swedish & 7 & 35.0 / \textbf{28.2}  &  \textbf{34.5} / 28.3 \\
 Japanese & 40 & 19.2 / 14.9 & \textbf{18.1} / \textbf{14.2}\\
 Chinese & 100 & 44.5 / 33.6 & 44.5 / \textbf{33.2} \\
 Mixtec & 0 & 67.4 / 59.7 &  \textbf{66.6} / \textbf{57.0}\\

\bottomrule
\end{tabular}

}
    \label{tab:mlang_cer}
    \vspace{-0.6cm}
\end{table}

All settings of WavLabLM achieve strong performance across the board, thanks to the multilingual training. WavLabLM MK outperforms WavLabLM EK on most tasks, particularly in the normal setting of the multilingual ASR (38.8 / 32.8 vs 41.0 / 33.5 CER) and ASR+LID tasks (38.6 / 32.8 vs 40.0 / 33.1 CER). This shows the importance of the initial feature extraction and cluster assignment step - multilingual k-means outperforms English k-means, even with the latter's additional pre-training on large-scale English data. However, reasonable performance can still be achieved with WavLabLM EK, as indicated by its higher SUPERB score on the 1-hour setting (862) vs WavLabLM MK (848). As such, better adaptation of monolingual models to multilingual settings may be a promising direction for future research, especially due to its computational efficiency (Table \ref{tab:models}).

Our multi-stage training technique (WavLabLM MS) was successful in better balancing the performance of the model, outperforming the unbalanced model (WavLabLM MK) in all tasks (908 / 897 vs 845 / 848 SUPERB$_s$) . To better understand the benefits of the technique, we break down the monolingual ASR results by language in Table \ref{tab:mlang_cer}. In both data settings, the multi-stage training generally degrades performance in languages that occupy a large portion of the pre-training data, such as English, French, and German. However, there are significant improvements in languages that are rarely seen during training, such as Japanese and Chinese. The multi-stage technique is particularly effective for generalizing to unseen languages during pre-training, reducing the CER on Mixtec by an absolute 0.8 and 2.7 CER on the 10-minute and 1-hour settings. Overall, WavLabLM MS achieves comparable performance to XLS-R 128 across many tasks. With less than 10\% of the pre-training data, it maintains 99\% and 91\% of the performance of XLS-R 128 on both the 10 minute and 1 hour tracks respectively (908 / 897 vs 917 / 986 SUPERB$_s$).

For our multi-resolution experiments, we found that CV-HuBERT 20ms was surprisingly powerful: it obtains similar performance to WavLabLM-MS on the 10 minute setting, and outperforms all WavLabLM variants on the 1 hour setting. CV-HuBERT 20m obtains 94\% of the performance of the SOTA XLS-R 128, with only 3\% of the training data and 30\% of the parameter size. While similar results of smaller models outperforming larger ones have been observed for monolingual SSL models \cite{shi2023ml,zaiem2023speech}, we are the first to observe such a trend in multilingual models. The light computational requirement of CV-HuBERT 20ms (about 1k GPU hours for 4 A100s) for its level of performance shows both the potential of SSL in academic settings, an exciting prospect for further research. 

Learning speech representations purely on lower resolutions significantly degraded performance, similar to the monolingual setting \cite{shi2023exploration}. Contrary to monolingual SSL, combining SSL models trained with multiple resolutions (CV-HuBERT MR) not only did not improve performance, but also degraded it in almost all tasks. This suggests that performant techniques for monolingual SSL may not translate well to multilingual settings, and shows the necessity of evaluating SOTA methods across a diverse set of linguistic settings.
\vspace{-0.2cm}
\subsection{Issues with Scaling}
\vspace{-0.2cm}
In our early experiments, we experimented with several modifications to the HuBERT architecture. This included using a convolutional-augmented encoder \cite{kimEbranch} and/or lightweight attention approximations \cite{dao2022flashattention}. However, we found that these changes significantly degraded training stability, which made scaling to larger models more difficult. As such, we opted to retain the vanilla HuBERT architecture with no modifications.

WavLM's utterance mixing technique requires each training batch to contain at least one audio clip. However, when dealing with large models, sometimes only a single clip can fit within a GPU. Our initial solution was to instead sample from the entire dataset by reading the sampled audio on-the-fly. However, this significantly slowed down training due to the additional file I/O. We also attempted to shard the optimizer across GPUs to reduce VRAM usage \cite{sharded} , but it resulted in increased runtime with little memory savings. Finally, we opted to instead approximate the WavLM-style augmentation by using only DNS noises in all batches with only one sample. Unfortunately, this still did not address the file I/O issue - the DNS noises were also read into memory on-the-fly. We ultimately decided on the computationally-expensive solution of always retaining all DNS noise recordings in memory, thus requiring the files to be read from disk only once.

\vspace{-0.3cm}
\section{Conclusion}
\vspace{-0.3cm}
We propose WavLabLM, a large-scale multilingual SSL model that extends the WavLM framework of joint prediction and denoising to 136 languages. We evaluate different pre-training methods for WavLabLM, focusing on pre-training targets and continual learning, in an effort to enhance the efficiency of SSL. In doing so, we devise a novel multi-stage pre-training approach, a lightweight solution designed to address the language imbalance of large-scale data. This led to our best model, WavLabLM MS, which maintains up to 99\% of the performance of SOTA multilingual SSL models on ML-SUPERB, with only 10\% of the pre-training data. We further slim down our method to create CV-HuBERT, which achieves similar performance with even less data. For future work, we hope to further scale WavLabLM to larger amounts of data, and apply it to weakly-supervised \cite{radford2022robust} settings to create even more powerful zero-shot models.\blfootnote{This work used Delta at NSCA through allocation CIS210014 from the ACCESS program, which is supported by NSF grants \#2138259, \#2138286, \#2138307, \#2137603, and \#2138296.}


\clearpage
\section{References}
\label{sec:ref}
{
\printbibliography
}
\end{document}